\documentclass[sigconf]{acmart}
\usepackage{multirow}
\usepackage{stfloats}
\pdfoutput=1
\AtBeginDocument{%
  }

\setcopyright{acmlicensed}
\copyrightyear{2018}
\acmYear{2018}
\acmDOI{XXXXXXX.XXXXXXX}
\acmConference[Conference acronym 'XX]{Make sure to enter the correct
  conference title from your rights confirmation email}{June 03--05,
  2018}{Woodstock, NY}
\acmISBN{978-1-4503-XXXX-X/2018/06}
\renewcommand\footnotetextcopyrightpermission[1]{}

\begin{document}

\title{LAD-Reasoner: Tiny Multimodal Models are Good Reasoners for Logical Anomaly Detection}

\author{Weijia Li}
\email{602024710006@smail.nju.edu.cn}
\affiliation{%
  \institution{Nanjing University}
  \city{Nanjing}
  \state{Jiangsu}
  \country{China}
}

\author{Guanglei Chu}
\email{chuguanglei@cmss.chinamobile.com}
\affiliation{%
  \institution{China Mobile (Suzhou) Software Technology Co., Ltd}
  \city{Suzhou}
  \state{Jiangsu}
  \country{China}
}

\author{Jiong Chen}
\email{chenjiong@cmss.chinamobile.com}
\affiliation{%
  \institution{China Mobile (Suzhou) Software Technology Co., Ltd}
  \city{Suzhou}
  \state{Jiangsu}
  \country{China}
}

\author{Guo-Sen Xie}
\email{gsxiehm@gmail.com}
\affiliation{%
 \institution{Nanjing University of Science and Technology}
 \city{Nanjing}
 \state{Jiangsu}
 \country{China}
}

\author{Caifeng Shan}
\email{cfshan@nju.edu.cn}
\affiliation{%
  \institution{Nanjing University}
  \city{Nanjing}
  \state{Jiangsu}
  \country{China}
}

\author{Fang Zhao}
\email{fzhao@nju.edu.cn}
\affiliation{%
  \institution{Nanjing University}
  \city{Nanjing}
  \state{Jiangsu}
  \country{China}
}


\begin{abstract}
Recent advances in industrial anomaly detection have highlighted the need for deeper logical anomaly analysis,
where unexpected relationships among objects, counts, and spatial configurations must be identified and explained.
Existing approaches often rely on large-scale external reasoning modules or elaborate pipeline designs,
hindering practical deployment and interpretability.
To address these limitations, we introduce a new task, Reasoning Logical Anomaly Detection (RLAD),
which extends traditional anomaly detection by incorporating logical reasoning.
We propose a new framework, LAD-Reasoner,
a customized tiny multimodal language model built on Qwen2.5-VL 3B.
Our approach leverages a two-stage training paradigm that first employs Supervised Fine-Tuning (SFT) for fine-grained visual understanding,
followed by Group Relative Policy Optimization (GRPO) to refine logical anomaly detection and enforce coherent, human-readable reasoning.
Crucially, reward signals are derived from both the detection accuracy and the structural quality of the outputs,
obviating the need for building chain of thought (CoT) reasoning data.
Experiments on the MVTec LOCO AD dataset show that LAD-Reasoner,
though significantly smaller,
matches the performance of Qwen2.5-VL-72B in accuracy and F1 score,
and further excels in producing concise and interpretable rationales.
This unified design reduces reliance on large models and complex pipelines,
while offering transparent and interpretable insights into logical anomaly detection.
Code and data will be released.
\end{abstract}




\settopmatter{printacmref=false}

\maketitle

\section{Introduction}
\footnote{This manuscript is a preprint and is currently under peer review.}
Industrial anomaly detection has traditionally focused on surface-level or appearance-based defects,
mainly about texture damage\cite{liu2024deep}.
Although significant progress has been made in detecting appearance-based anomalies in industrial applications\cite{gu2024filo,yang2024promptable,li2024promptad},
the domain of logical anomaly detection\cite{tong2025component,hsieh2024csad,yang2024slsg},
where anomalies stem from subtle relational inconsistencies,
has received limited attention.
Addressing such logical anomalies requires models not only to detect deviations accurately but also to explain the underlying reasoning in a way that is transparent to humans.

\begin{figure}[t]
  \centering
  \includegraphics[width=\linewidth]{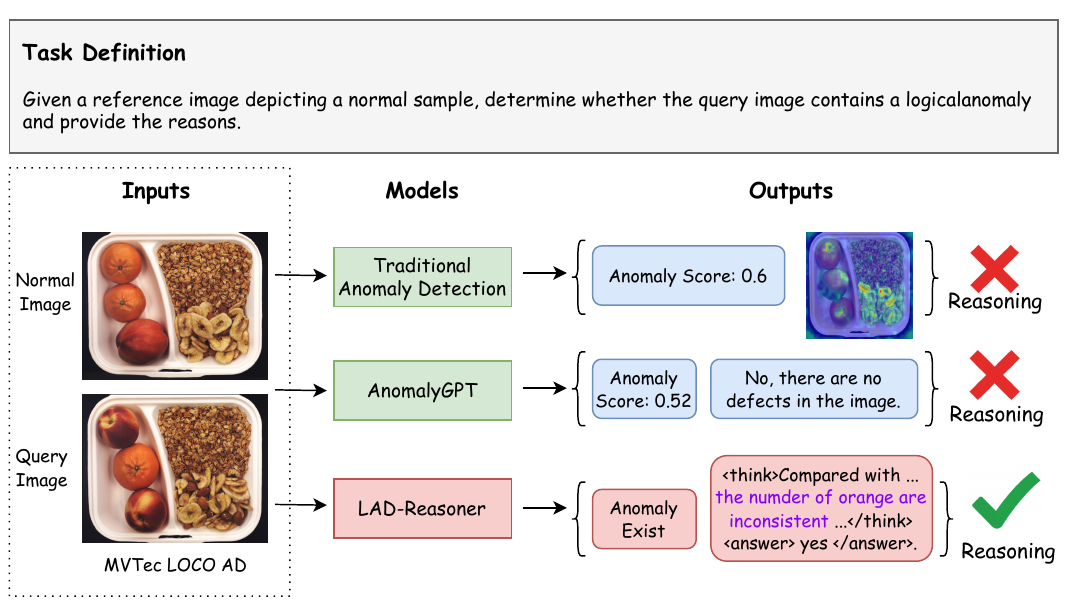}
  \caption{Overview of the task definition and comparison among existing traditional methods, MLLM-based methods, and our proposed LAD-Reasoner. While prior approaches fail to provide human-interpretable reasoning for anomaly detection, LAD-Reasoner delivers both accurate predictions and readable reasoning process.}
  \Description{Comparison with other methods.}
  \label{fig:The first image}
\end{figure}
%
Recent advances in large-scale reasoning models (e.g., OpenAI-o1\cite{jaech2024openai} and DeepSeek-R1\cite{guo2025deepseek}) have demonstrated strong contextual and relational inference capabilities in test-time scaling (TTS),
that aims to increase the compute at test time to get better results,
yet they were not initially designed for industrial logical anomaly detection.
Currently, high-performing methods for logical anomaly detection are training-free\cite{jin2025logicad,zhang2024logicode,zhang2025towards},
and often rely on intricate pipelines that integrate specialized modules and incorporate powerful language models like GPT-4 series model\cite{hurst2024gpt,yang2023dawn}
to generate reasoning chains.
Nevertheless, their reasoning processes tend to be opaque,
limiting human readability and thus hindering broader practical adoption.

Meanwhile,
reinforcement learning methods\cite{zhang2025mm,rafailov2023direct,yu2022surprising} have shown effectiveness in policy optimization.
For instance,
Group Relative Policy Optimization (GRPO)\cite{shao2024deepseekmath} enhances learning efficiency by leveraging groupwise reward comparisons.
GRPO eschews the need for a separate critic network, thereby reducing computational overhead and stabilizing training by diminishing the variance of policy gradients.
It also allows for greater controllability through KL divergence constraints, preventing drastic policy updates.
Notably, although these characteristics align well with the requirements of logical anomaly detection,
especially for generating human-readable reasoning,
existing work on logical AD has not fully leveraged GRPO's potential to simplify architecture design and improve the explainability of results.
Motivated by these developments and the gap in existing logical AD research,
we propose a new task, Reasoning Logical Anomaly Detection (RLAD),
which aims to detect logical anomalies while generating human-interpretable explanations of those anomalies.
To tackle this task,
we introduce LAD-Reasoner,
a customized tiny multimodal language model based on Qwen2.5-VL 3B\cite{bai2025qwen2}.
It adopts a two-stage training framework that combines Supervised Fine-Tuning (SFT) and GRPO,
enabling interpretable reasoning for logical anomaly detection,
and its abilities are shown in Figure~\ref{fig:The first image}.
In the SFT stage,
the model is equipped with fine-grained visual understanding,
trained on approximately 3k image-caption pairs sourced from the test sets of MVTec AD\cite{bergmann2019mvtec} and VisA\cite{zou2022spot}.
Subsequently,
the GRPO stage uses rule-based reward functions over a lightweight dataset of no more than 1k image pairs,
where each pair includes a reference normal image and a test image labeled as anomalous or not.
These image pairs are also selected from the test sets of MVTec AD and VisA,
specifically curated to include samples with logical and functional anomalies.
The model receives reward signals purely from prediction correctness,
avoiding the need for hand-crafted reasoning annotations, alongside structure rewards that ensure structured,
coherent reasoning outputs.
By unifying SFT for visual perception with GRPO for reward-driven optimization,
our framework achieves robust anomaly detection and generates interpretable reasoning statements on the MVTec LOCO AD dataset,
without relying on excessively large models or complex multi-stage pipelines.

Our key contributions are summarized as follows:
\begin{enumerate}
  \item \textbf{Unified Design}: We present a unified two-stage framework integrating SFT and GRPO, eliminating complex pipelines while remaining easy to follow.

  \item \textbf{Data and Model Efficiency}: Our method requires only a fraction of the data and computational resources, yet achieves competitive performance compared to existing methods, even Qwen2.5-VL 72B, for logical anomaly detection.

  \item \textbf{Enhanced Interpretability}: By generating natural language reasoning, our framework offers transparent insights into the anomaly detection process, facilitating easier validation and adoption in industrial settings.
\end{enumerate}

\section{Related Work}
\subsection{Logical Anomaly Detection}
Logical Anomaly Detection (Logical AD) focuses on identifying violations of logical constraints and relational inconsistencies within structured data\cite{guo2023template}.
While appearance anomaly detection has received considerable attention\cite{zhou2023anomalyclip,cao2024adaclip}, logical anomalies remain relatively underexplored.
Recent approaches\cite{tong2025component,hsieh2024csad,yang2024slsg} to logical anomaly detection have focused on identifying inconsistencies in object relationships and contextual arrangements.
LogSAD\cite{tong2025component} multimodal anomaly detection framework leverages GPT-4V to generate matching rules based on visual-textual alignment.
It employs multi-granularity anomaly detectors to capture anomalies from various perspectives,
and integrates calibrated scores to effectively detect logical anomalies.
LogiCode\cite{zhang2024logicode} introduces a framework that leverages large language models to extract logical rules
from normal images and generate corresponding Python code for detecting logical anomalies in industrial scenes.
LogicAD\cite{jin2025logicad} leverages advanced vision-language models (AVLMs) to extract rich textual features.
By incorporating guided chain-of-thought reasoning, region-of-interest segmentation,
and text embedding filtering,
it computes anomaly scores through format-aware embeddings and generates interpretable logical inferences by automated theorem prover.
However, these training-free methods often rely on complex framework designs
and the integration of external models such as the GPT-4 series,
which hinders seamless end-to-end deployment.
Furthermore,
the generation of human-interpretable rationales for detected logical anomalies remains a largely unaddressed problem.
These limitations highlight the need for compact, interpretable, and end-to-end frameworks that can unify detection and reasoning without relying on large-scale external models.

\subsection{Multimodal Large Language Models}
Current approaches\cite{yao2024minicpm,liu2023visual,bai2025qwen2} have increasingly extended Large Language Models (LLMs) to handle multimodal inputs,
including visual and auditory information, moving beyond their original text-centric design.
This evolution has broadened their application domains and opened new avenues for addressing complex tasks in industrial anomaly detection\cite{gu2024anomalygpt,li2023myriad,xu2025towards}.
%
There are some MLLM for anomaly detection,
%
AnomalyGPT\cite{gu2024anomalygpt} leverages the MiniGPT4\cite{zhu2023minigpt} architecture to perform few-shot anomaly detection. While this approach demonstrates promising results,
its fixed framework design limits its flexibility in adapting to varied user instructions.
In contrast, Myriad\cite{li2023myriad} adopts a different strategy by incorporating existing industrial anomaly detection as specialized visual experts to guide the large model,
thereby effectively bridging the gap between textual and visual domains.
Another noteworthy method, Anomaly-OV\cite{xu2025towards},
achieves state-of-the-art performance by utilizing a comprehensive 125k instruction tuning dataset.
Despite its robust anomaly detection capabilities,
Anomaly-OV pays limited attention to logical anomaly detection,
particularly those requiring an understanding of complex object relationships and structured arrangements.
Overall, although current multimodal frameworks have advanced the field of anomaly detection,
they still fall short in integrating robust logical reasoning capabilities.
The key challenge is to effectively combine deep learning-based multimodal processing with formal logical reasoning to handle anomalies that arise from intricate relational structures. 
Motivated by these limitations,
our work focuses on enhancing the reasoning abilities of multimodal models specifically for anomaly detection tasks that involve complex logical constraints.
%

\subsection{Logical Reasoning}
The evolution of reasoning in Large Language Models progressed from basic in-context learning to Chain of Thought (CoT)\cite{wei2022chain} prompting,
which significantly improved performance on tasks requiring logical reasoning by breaking problems into intermediate steps.
Notable advances include OpenAI's o1 models\cite{jaech2024openai},
which demonstrated impressive reasoning abilities for complex problems,
and Deepseek-R1\cite{guo2025deepseek}, which regard Group Relative Policy Optimization (GRPO)\cite{shao2024deepseekmath} as an effective approach for enhancing reasoning capabilities.
Although GRPO has been successfully integrated into many domains to boost performance by leveraging spontaneous reasoning\cite{liu2025seg,zhang2025r1,huang2025vision},
industrial applications still predominantly rely on prompted CoT reasoning,
thereby limiting the effectiveness in detecting logical anomalies in complex scenarios.

\section{Preliminary}
GRPO consists of a policy model $\pi_\theta(\cdot)$ and reward model $r_\phi(\cdot)$, both based on pre-trained LLMs.
The core innovation is using group sampling to estimate advantages without a value function.
For a question $q$, GRPO samples multiple outputs $\{o_1, o_2, \ldots, o_G\}$ from policy $\pi_{\theta_{\text{old}}}$
to obtain reward distribution $\{r_1, r_2, \ldots, r_G\}$.
The normalized reward $\tilde{r}_i = \frac{r_i-\text{group-mean}(r)}{\text{group-std}(r)}$ serves as the advantage for all tokens in that output. GRPO maximizes:
\begin{align}
J_{\text{GRPO}}(\theta) &= \mathbb{E}\Bigg[\frac{1}{G} \sum_{i=1}^G \frac{1}{|o_i|} \sum_{t=1}^{|o_i|} \Big[\min\big(\rho_{i,t}\hat{A}_{i,t},\, \nonumber \\
&\quad \text{clip}(\rho_{i,t},\, 1-\varepsilon,\, 1+\varepsilon)\hat{A}_{i,t}\big)\Big] - \beta D_{\text{KL}}\Bigg],
\end{align}
\noindent where $D_{\text{KL}} = D_{\text{KL}}[\pi_\theta||\pi_{\text{ref}}]$ and $\rho_{i,t} = \frac{\pi_\theta(o_{i,t}|q,o_{i,<t})}{\pi_{\theta_{\text{old}}}(o_{i,t}|q,o_{i,<t})}$.
Here, $\varepsilon$ and $\beta$ are hyperparameters that control clipping threshold and KL penalty strength, respectively.
Unlike traditional PPO\cite{yu2022surprising} that requires a separate value network,
GRPO efficiently uses group statistics of sampled outputs as the baseline for advantage calculation,
significantly reducing computational requirements while maintaining performance on complex reasoning tasks.

\section{Approach}

\subsection{Data Generation}

\subsubsection{Data for SFT}
SFT is essential for enhancing model performance and domain-specific understanding,
yet collecting sufficient high-quality data remains a major challenge,
especially for both appearance and logical anomalies in industrial contexts.
Although MMAD\cite{jiangmmad} provides a sizable set of anomaly-related annotations,
they are not directly suitable for SFT due to their annotation format.

\begin{figure}[t]
  \centering
  \includegraphics[width=\linewidth]{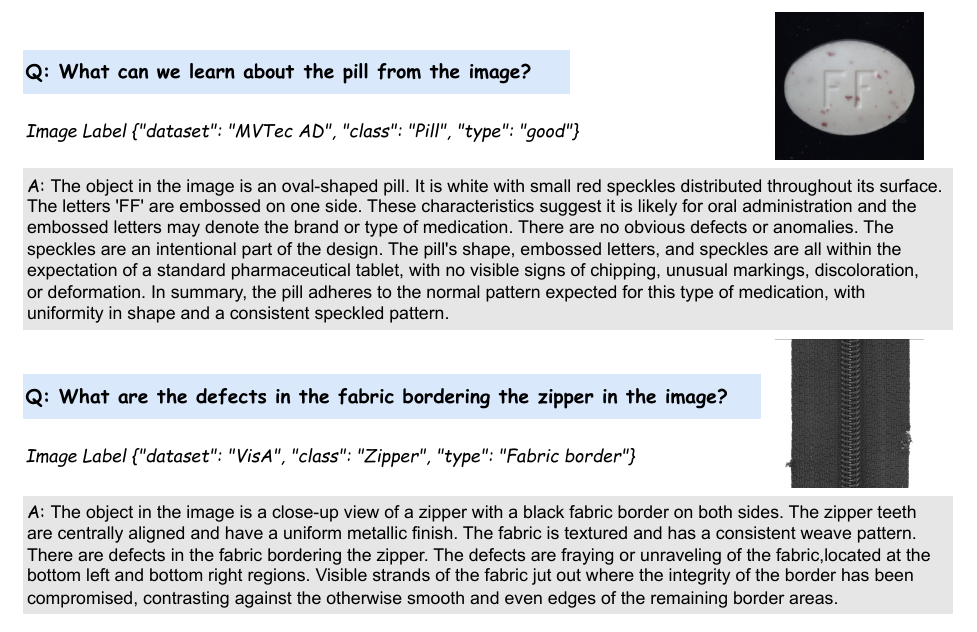}
  \caption{Examples of input–output pairs used for SFT. Each sample consists of a question as a prompt, an image, and a corresponding answer mainly describing about anomaly.}
  \Description{SFT Data Show.}
  \label{fig:The data show image}
\end{figure}

To address this, we construct an SFT dataset by repurposing test samples from MVTec and VISA, focusing on appearance-level defects.
This ensures strong baseline quality and helps the model capture fine-grained visual cues.
To further enrich data contents while maintaining relevance,
we design a generation pipeline that strategically leverages MMAD annotations as semantic priors.
These annotations offer concise descriptions of known anomalies and contextual cues,
serving as reliable anchors for guided generation.
Instead of serving as ground-truth labels,
they guide the model where to focus,
helping it explore finer details without drifting from the anomaly semantics.
For each image,
we craft a structured prompt for Gemini-2.0-Pro-Exp\cite{Gemini2025},
specifying artifact type and referencing the MMAD information.
We adopt Gemini for data generation due to its competitive visual understanding capabilities and significantly lower computational cost compared to models like GPT-4o.
The model is instructed to examine the image beyond the given priors and generate detailed QA pairs that combine both referenced and newly discovered content.
As shown in Figure~\ref{fig:The data show image},
these examples highlight fine-grained image details,
thereby enhancing the model's capability to capture fine details.
%
%
To ensure consistency and correctness across generations, we periodically sample outputs for manual review.
This process yields 2,731 high-quality QA samples,
demonstrating that, when properly leveraged, external annotations can significantly reduce labeling costs while enriching data quality and semantic precision.

\subsubsection{Data for GRPO}
Traditional reinforcement learning methods for anomaly detection typically require reasoning-oriented data with rich annotations.
In contrast,
our GRPO training leverages a lightweight and easily constructible dataset designed to enable comparative reasoning.
Specifically, we select several MVTec classes that exhibit subtle,
context-dependent anomalies resembling logical inconsistencies.
For each test image,
we pair it with a normal training image from the same class to form a reference-query input.
In our logical anomaly detection tasks, we deliberately simplify the prompt provided to the MLLM by using \textit{anomaly} instead of \textit{logical anomaly}.
This approach is intended to prevent potential misinterpretations by smaller-scale models.
Moreover, the training data for GRPO primarily cover number,
color, existence, and functional anomalies,
where the notion of anomaly is sufficiently representative.
Specifically, we set the prompt as:
we set the prompt directly: \textit{Can you find any anomaly in the query image compared to the reference?}
A unified prompt guides the model to determine whether the query is anomalous with respect to the reference.
Binary supervision is then assigned based on the test label.
This simple yet effective construction yields approximately 984 QA samples,
significantly reducing the complexity of data preparation compared
\begin{figure*}[t]
  \centering
  \includegraphics[width=\linewidth]{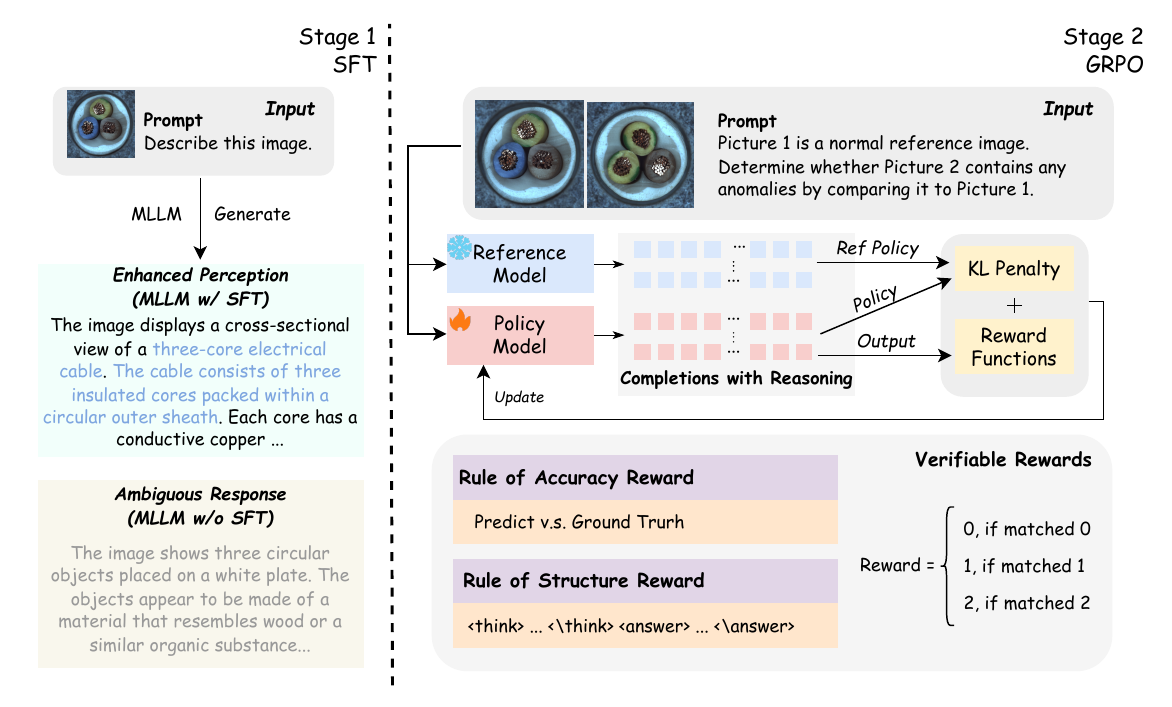}
  \caption{The architecture of LAD-Reasoner. The training process consists of two stages. In the first stage, applying SFT to the base MLLM leads to improved visual detail understanding. In the second stage, the policy model is optimized based on verified rewards and the KL divergence penalty, enabling it to generate outputs that conform to a predefined structure and yield accurate final predictions..}
  \Description{Comparison with other methods.}
  \label{fig:The second image}
\end{figure*}
to conventional reinforcement learning approaches,
while still enabling the model to acquire essential reasoning abilities during GRPO training.
We aim for the model to learn to identify and reason about logical anomalies—including those of logical and functional nature—through this dataset.

\subsection{LAD-Reasoner}
In this section,
we introduce LAD-Reasoner,
a unified framework built upon Qwen2.5-VL 3B that synergistically combines SFT and GRPO,
to enhance the performance of lightweight multimodal models on logical anomaly detection tasks.
The SFT stage focuses on appearance-level anomalies.
By fine-tuning the model on a dataset enriched with fine-grained artifact descriptions,
we improve its sensitivity to subtle yet critical visual cues,
allowing it to better capture anomaly-related details that are often overlooked by general-purpose models,
as shown in the left part ot Figure~\ref{fig:The second image}.
As shown in it, the model without SFT produces vague and semantically incorrect descriptions (e.g., referring to an electrical cable as wooden objects).
In contrast,
the fine-tuned model generates accurate,
structured descriptions that align closely with the visual content,
demonstrating improved detail perception critical for logical anomaly understanding.
Complementing SFT,
the GRPO stage aims to strengthen reasoning ability of the model in detecting logical inconsistencies.
The GRPO training process involves four main steps.

\textbf{(1)Prompt Sampling and Response Generation:}
In each training iteration,
a batch of prompts is sampled from the training corpus and fed into both the policy model and a frozen reference model to generate multiple candidate responses in parallel.
The policy model produces diverse outputs under structured decoding constraints guided by a predefined reasoning template,
ensuring consistency and interpretability across trajectories.
We guide the model to output a structured reasoning trace using a predefined template:
\textit{\textless think\textgreater...\textless /think\textgreater},
followed by a binary decision enclosed in \textit{\textless answer\textgreater \ ... \textless /answer\textgreater} tags.
i.e., \textit{\textless think\textgreater\ think process \textless /think\textgreater\textless answer\textgreater\ yes or no \textless /answer\textgreater}.
Rather than naively combining different response components,
our approach integrates structured prompting with diverse response sampling to promote both compositionality and semantic alignment.
This strategy ensures that generated reasoning traces remain coherent and grounded,
which is critical for downstream reward computing.

\textbf{(2)Log-Probability Computation:}
We compute token-level log-probabilities for each output sequence under both the trainable policy model and a frozen reference model.
Given an input query $q$, the trainable policy model $\pi_\theta$ generates multiple candidate output sequences,
forming a sampled group.
Each output sequence is denoted as $o_i = (a_{i,1}, a_{i,2}, \ldots, a_{i,|o_i|})$,
where $a_{i,t}$ is the $t$-th token of the $i$-th sampled sequence.
To support fine-grained reward modeling,
we compute token-level log-probabilities of each token under both the current policy and a frozen reference model $\pi_{\text{ref}}$:
\begin{equation}
  \ell_{\theta}(a_{i,t} \mid q, a_{i,<t}) = \log \pi_{\theta}(a_{i,t} \mid q, a_{i,<t}),
\end{equation}
\begin{equation}
  \ell_{\text{ref}}(a_{i,t} \mid q, a_{i,<t}) = \log \pi_{\text{ref}}(a_{i,t} \mid q, a_{i,<t}).
\end{equation}
To further support sequence-level reward estimation and policy regularization,
we aggregate the token-level log-probabilities to compute the total log-probability of each sequence:
\begin{equation}
  \ell_{\theta}(o_i \mid q) = \sum_{t=1}^{|o_i|} \log \pi_{\theta}(a_{i,t} \mid q, a_{i,<t}),
\end{equation}
\begin{equation}
  \ell_{\text{ref}}(o_i \mid q) = \sum_{t=1}^{|o_i|} \log \pi_{\text{ref}}(a_{i,t} \mid q, a_{i,<t}).
\end{equation}
Based on these, we define the KL penalty as the difference in sequence-level log-probabilities:
\begin{equation}
  \mathcal{R}_{\text{KL}}(o_i) = \ell_{\theta}(o_i \mid q) - \ell_{\text{ref}}(o_i \mid q).
\end{equation}
These token-level log-probabilities not only reflect the model’s confidence at each step of generation,
but also serve as the foundation for subsequent \textit{reward computation} and \textit{advantage estimation}.
By comparing the probabilities under $\pi_\theta$ and $\pi_{\text{ref}}$,
we derive token-level importance weights that capture distributional shifts in the policy’s behavior.
These weights play a central role in guiding stable and targeted policy updates under the GRPO framework.

\textbf{(3)Reward Computation and Advantage Estimation:}
To optimize the policy,
we construct a scalar reward signal composed of verifiable supervision and a regularization term.
Specifically,
each output sequence \( o_i \) is assigned a composite reward:
\begin{equation}
\mathcal{R}(o_i) = \lambda_{\text{struct}} \cdot \mathcal{R}_{\text{struct}}(o_i) + \lambda_{\text{acc}} \cdot \mathcal{R}_{\text{acc}}(o_i) - \beta \cdot \mathcal{R}_{\text{KL}}(o_i).
\end{equation}
Here, \( \mathcal{R}_{\text{struct}} \) denotes the \textit{Structure Reward},
which encourages the model to generate coherent reasoning traces strictly within the \texttt{<think>}...\texttt{</think>} tags.
It is computed as a binary score based on whether the reasoning content adheres to the predefined format.
\(\mathcal{R}_{\text{acc}}\) denotes the \textit{Accuracy Reward},
which validates whether the final binary decision (enclosed in \texttt{<answer>}...\texttt{</answer>}) matches the ground-truth label.
The reward is set to 1 if and only if the predicted answer exactly matches the ground-truth label (i.e., both are either \texttt{yes} or \texttt{no});
otherwise, the reward is 0.
For example, if the ground-truth label is \texttt{yes},
the model must output \texttt{<answer> yes </answer>} to obtain a reward of 1;
any other output results in a reward of 0.
\( \mathcal{R}_{\text{KL}} \) is the KL penalty previously defined in the previous part, constraining the policy to remain close to the reference model and stabilizing training dynamics.
As only the final token in each response receives a non-zero reward, we treat the trajectory-level reward as a terminal reward signal.
The advantage is computed at the sequence level using a simple Monte Carlo estimator:
\begin{equation}
A(o_i) = \mathcal{R}(o_i) - b(q).
\end{equation}
Unlike token-level methods such as GAE\cite{schulman2015high},
we do not propagate intermediate rewards across tokens,
as only the final prediction contributes to the reward.

\textbf{(4) Structured Reasoning Trace and Policy Update}  
Given the computed scalar rewards and advantages, the policy model is updated via minibatch optimization with gradient accumulation. We adopt a reinforcement learning objective that integrates both the advantage signal and a structured reasoning constraint. The overall objective is:
\begin{equation}
\mathcal{L}_{\text{GRPO}} = -\mathbb{E}_{o_i \sim \pi_\theta} \left[ w_i \cdot A(o_i) \right],
\end{equation}
where \( A(o_i) \) is the trajectory-level advantage as defined earlier,
and \( w_i \) denotes the group-relative importance weight defined by GRPO.
Structured reasoning traces, enclosed in \texttt{<think>}...\texttt{</think>} and \texttt{<answer>}...\texttt{</answer>} tags,
are enforced via structured prompting and decoding constraints.
These not only serve as format priors during generation,
but also guide the policy to align with verifiable logic structures.

By reinforcing both the logical coherence of intermediate reasoning and the factual correctness of final decisions,
our policy optimization approach yields models that are not only more accurate but also significantly more interpretable.
This structured,
reward-guided training paradigm leads to better semantic alignment and robustness in anomaly reasoning,
offering a scalable pathway toward reasoning logical anomaly detection.

\begin{table*}[b]
\caption{
Overall and per-category results (Accuracy / F1) of different approaches on MVTec LOCO AD. \textbf{Overall} evaluation over the test set without per-class averaging. JB, BB, SB, SC, and PP denote Juice Bottle, Breakfast Box, Screw Bag, Splicing Connectors, and Pushpins, respectively. \textbf{Bold} and \underline{underline} indicate the best and second-best scores.
}
\label{tab:results}
\centering
\begin{tabular}{lcccccccccccc}
\toprule
\multirow{2}{*}{\textbf{Model}} &
  \multicolumn{2}{c}{\textbf{Overall}} &
  \multicolumn{2}{c}{\textbf{JB}} &
  \multicolumn{2}{c}{\textbf{BB}} &
  \multicolumn{2}{c}{\textbf{SB}} &
  \multicolumn{2}{c}{\textbf{SC}} &
  \multicolumn{2}{c}{\textbf{PP}} \\
 &
  \textbf{Acc.} &
  \textbf{F1} &
  \textbf{Acc.} &
  \textbf{F1} &
  \textbf{Acc.} &
  \textbf{F1} &
  \textbf{Acc.} &
  \textbf{F1} &
  \textbf{Acc.} &
  \textbf{F1} &
  \textbf{Acc.} &
  \textbf{F1} \\ \midrule
APRIL-GAN &
  52.4 &
  47.1 &
  50 &
  55.3 &
  64.3 &
  35.3 &
  52.9 &
  50.4 &
  \textbf{59.5} &
  45.9 &
  35.4 &
  \underline{48.6} \\
AnomalyGPT &
  42.6 &
  57.3 &
  50.4 &
  65.9 &
  39.5 &
  56.3 &
  39 &
  51.2 &
  46.3 &
  \underline{59.6} &
  38.0 &
  \textbf{53.3} \\ \midrule
Qwen2.5-VL-3B &
  58.8 &
  54.8 &
  60.2 &
  \underline{67.1} &
  74.1 &
  63.6 &
  47.5 &
  39.8 &
  52.9 &
  61.1 &
  \textbf{63.8} &
  29.1 \\
Qwen2.5-VL-72B &
  \textbf{62.1} &
  \underline{59.6} &
  \textbf{64.4} &
  66.7 &
  \underline{74.6} &
  \textbf{73.7} &
  \textbf{54.8} &
  \underline{60.6} &
  \underline{57.3} &
  59.1 &
  \underline{62.9} &
  14.1 \\ \midrule
Qwen2.5-VL-3B-SFT &
  46.3 &
  43.5 &
  49.2 &
  56.5 &
  29.2 &
  15.5 &
  44.0 &
  52.8 &
  44.5 &
  46.6 &
  61.6 &
  17.0 \\
Qwen2.5-VL-3B-GRPO &
  58.1 &
  55.6 &
  56.4 &
  65.3 &
  71.4 &
  53.1 &
  49.4 &
  53.0 &
  52.4 &
  \textbf{61.7} &
  64.6 &
  19.8 \\
LAD-Reasoner (ours) &
  \underline{60.4} &
  \textbf{63.5} &
  \underline{62.7} &
  \textbf{74.7} &
  \textbf{76.8} &
  \underline{68.6} &
  \underline{53.3} &
  \textbf{66.3} &
  52.9 &
  56.3 &
  60.3 &
  37.2 \\ \bottomrule
\end{tabular}
\end{table*}

\section{Experiments}

\subsection{Evaluation Settings}
To ensure a fair and meaningful evaluation,
we compare our method with approaches specifically designed for scenarios involving both zero-shot and few-shot constraints,
where the models are trained without access to the target dataset and evaluated in a one-shot mode on unseen data.
Hence,
APRIL-GAN\cite{chen2023zero} and AnomalyGPT\cite{gu2024anomalygpt} are chosen as representative methods for comparison. The former is evaluated via binary classification using a 0.5 threshold on predicted scores,
while the latter is assessed based on its generated textual outputs.
In addition, we introduce two ablated variants of our method to isolate the impact of each training stage: one using only SFT and another using only GRPO, without combining the two.
Notably, these methods are evaluated on, but not trained with, MVTec LOCO AD.
We adopt accuracy and F1-score as our main evaluation metrics.
Accuracy measures the overall prediction correctness,
while the F1-score offers a more balanced and informative assessment of model performance.

\subsection{Implementation Details}
Our training pipeline consists of two stages,
that are SFT and GRPO.
In the SFT stage,
we utilize the LLaMAFactory framework\cite{zheng2024llamafactory} for visual instruction tuning.
The model is trained via full-parameter fine-tuning on our collected SFT dataset for 3 epochs with a learning rate of 1e-6 and a batch size of 16.
For the GRPO stage,
we adopt EasyR1\cite{sheng2024hybridflow,zheng2025easyr1} as the training framework.
The model is trained on our custom-designed GRPO toy dataset for 50 steps using the same learning rate (1e-6) and a batch size of 64.
$\lambda_{\text{struct}}$ and $\lambda_{\text{acc}}$ are set to 0.5 for balance.

\subsection{Results and Analysis}
As shown in Table~\ref{tab:results},
Our proposed method, LAD-Reasoner,
demonstrates impressive performance in both Accuracy and F1 score,
showcasing strong generalization across diverse categories.
Notably,
it delivers competitive results compared with Qwen2.5-VL-72B,
while generally outperforming previous anomaly detection approaches.
In this task, LAD-Reasoner significantly outperforms APRIL-GAN and AnomalyGPT,
benefiting from its visual detail perception and interpretable reasoning capabilities.
However, in Pushpins cases,
the model's performance drops unexpectedly,
revealing challenges in handling fine-grained spatial anomalies.
We analyze this failure case to better understand the limitations and potential improvements.
Following this, we further analyze the qualitative outputs generated by LAD-Reasoner.

\begin{figure*}[t]
  \centering
  \includegraphics[width=\linewidth]{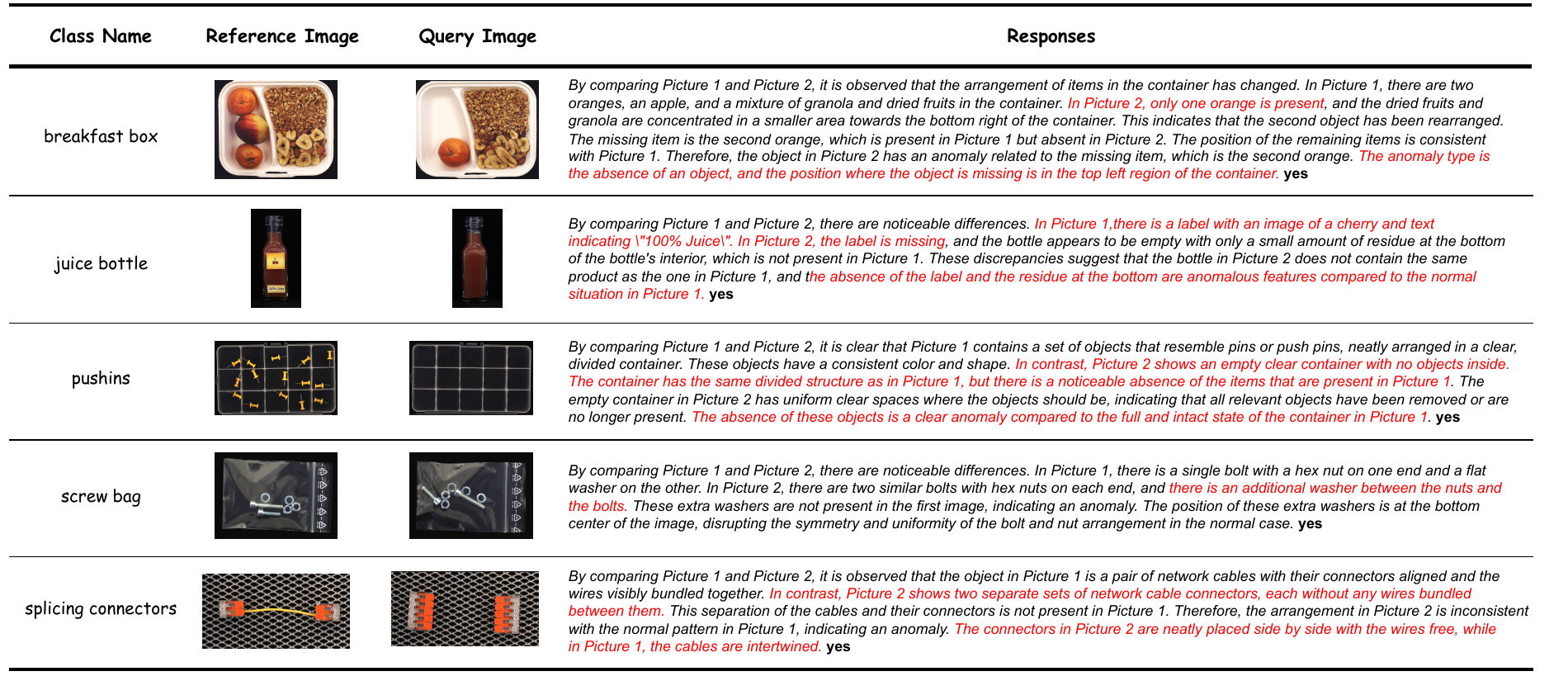}
  \caption{Visualization of the inference results produced by LAD-Reasoner. For each subclass in the MVTec LOCO AD dataset, a representative test case is presented, including a reference image, a query image, and a natural language prompt inquiring whether an anomaly is present. The model responds with a thinking process (shown in \textit{italic}) followed by a binary decision (shown in \textbf{bold}). For clarity of presentation, the original \texttt{<think>}\texttt{<\textbackslash{}think>} and \texttt{<answer>}\texttt{<\textbackslash{}answer>} tags are omitted.}
  \Description{Results Visualization.}
  \label{fig:The third image}
\end{figure*}

As shown in Figure~\ref{fig:The third image}, it illustrates qualitative examples on five sub-classes from the MVTec LOCO AD dataset.
Each row compares a reference image (normal) and a query image (possibly anomalous),
along with the response generated by our LAD-Reasoner model.
Reasoning Logical Anomaly Detection(RLAD) involves subtle semantic or logical inconsistencies that are difficult to depict by simple pixel-level differences.
The most notable feature of LAD-Reasoner is its ability to generate detailed reasoning paths in natural language.

As shown in the responses of models,
the italicized text segments represent the model’s intermediate reasoning steps,
demonstrating its fine-grained visual perception and logical reasoning capabilities in test-time scaling (tts).
In the breakfast box case, juice bottle, and screw bag cases,
LAD-Reasoner is able to spontaneously reason about anomalies such as the number of fruits,
the absence of product labels,
and the presence or absence of specific components, without being guided by handcrafted prompts.
These observations show that the model can identify meaningful differences grounded in semantic understanding rather than superficial visual changes.
Furthermore, in the pushpins case,
the model accurately recognizes the spatial structure of the reference container and identifies fine-grained geometric inconsistencies.
It not only detects the absence of objects in the designated compartments but also demonstrates an awareness of numerical discrepancies,
highlighting its ability to perceive both structural and quantitative anomalies.
Notably, LAD-Reasoner is also capable of identifying functional anomalies,
as seen in the splicing connectors example,
where the deviation in connector positioning may impact the connectivity of the network cable.
This highlights the model’s potential for practical deployment in scenarios requiring both visual and functional reasoning.
These reasoning capabilities are acquired during the GRPO stage,
where the model learns to perform structured and visually grounded reasoning without relying on task-specific instructions or handcrafted prompts.
The resulting long-form explanations not only lead to accurate anomaly detection,
but also enhance the interpretability of the decision process by providing human-understandable reasoning aligned with visual semantics.
This highlights the robustness and generalizability of LAD-Reasoner in practical, open-ended anomaly detection scenarios.

Moreover, the reasoning process demonstrates that even a 3B-scale model possesses certain reasoning abilities, and the answers are generated in the expected structured format.

\section{Discussion}

\subsection{The Effect of SFT.}
\begin{figure}[b]
  \centering
  \includegraphics[width=\linewidth]{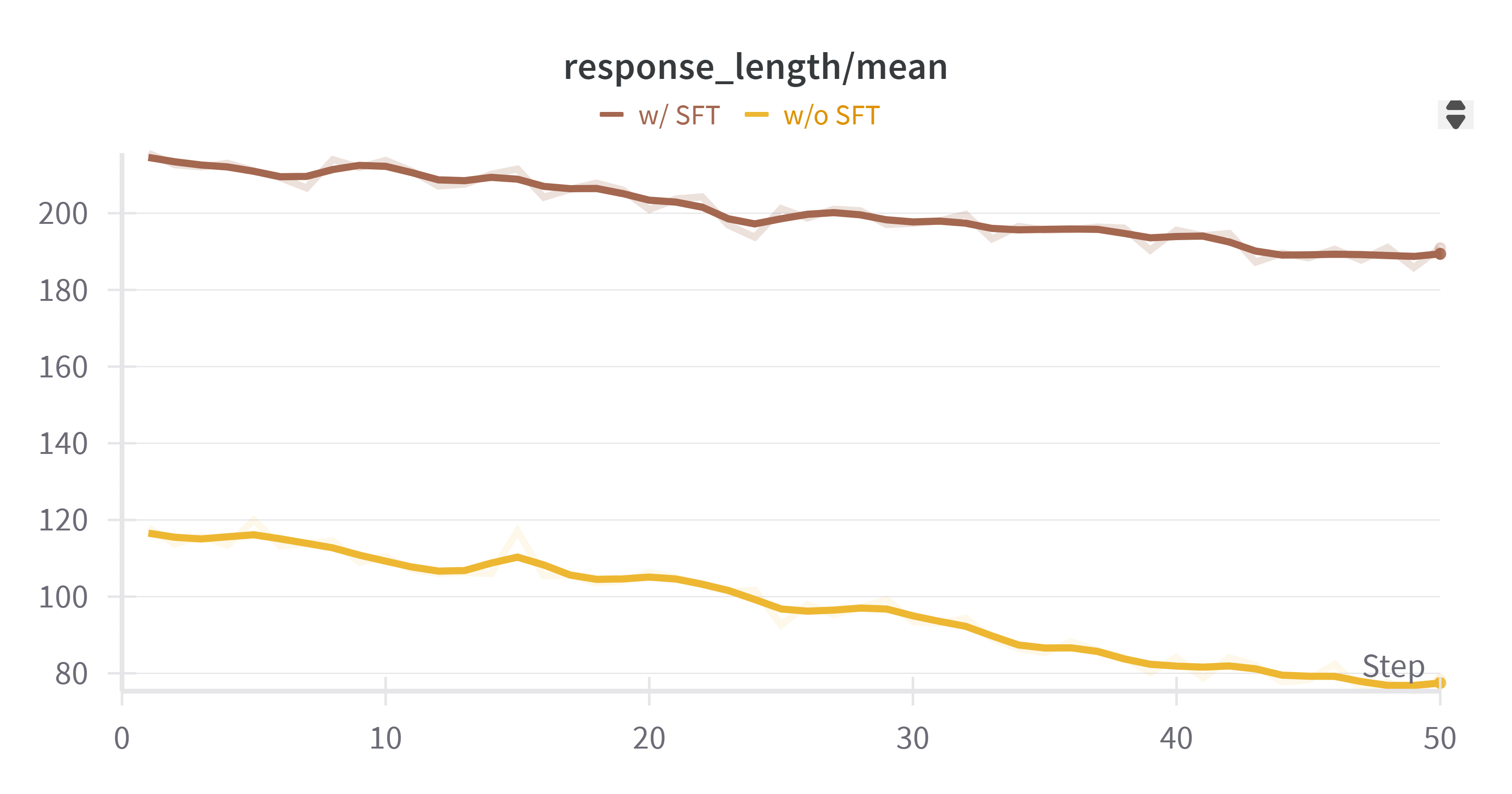}
  \caption{The response lengths of models with and without SFT during the GRPO stage. The upper curve corresponds to the model with SFT, while the lower curve represents the model without SFT.}
  \Description{Comparison with other methods.}
  \label{fig:length}
\end{figure}
As shown in Table~\ref{tab:results},
models trained with Supervised Fine-Tuning (SFT) sometimes exhibit slightly lower final performance in our RLAD task.
A similar phenomenon has also been observed in LLaMA 4\cite{meta2024llama4},
where SFT was found to constrain the exploration behavior during the reasoning process.
This is because SFT tends to encourage the model to follow fixed response patterns, whereas our task requires the model to engage in open-ended reasoning.

In our experiments,
relying solely on SFT limits the model’s ability to explore diverse reasoning paths,
which are essential for logical anomaly detection.
However,
SFT remains crucial for establishing the model’s grounding in visual semantics capability,
where SFT enhances the model's perceptual sensitivity to fine-grained details.
The influence of SFT also manifests in the model's response length during the GRPO stage and final inference.
As shown in Figure~\ref{fig:length},
the model trained with SFT consistently generates longer responses compared to its non-SFT counterpart during the GRPO stage.
This suggests that SFT equips the model with richer knowledge and more structured reasoning patterns,
enabling LAD-Reasoner to achieve better performance when combined with GRPO.
Overall, while SFT may slightly limit exploratory behavior,
it plays an essential role in initializing the model’s perception and reasoning capabilities,
which are further refined through reinforcement like GRPO.

\subsection{Efficiency and Performance Trade-offs}
\begin{figure}[t]
  \centering
  \includegraphics[width=\linewidth]{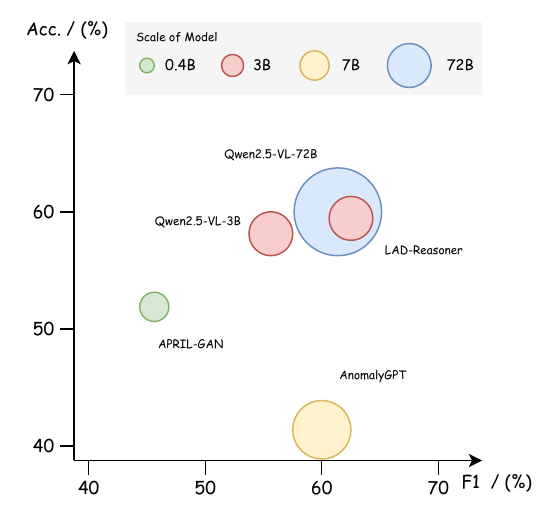}
  \caption{Comparison of model performance in terms of Accuracy and F1 score. Bubble size indicates the number of parameters.}
  \Description{Para Show.}
  \label{fig:param}
\end{figure}
Figure~\ref{fig:param} presents a comparative analysis of various models in terms of accuracy and F1 score,
with bubble size indicating model scale (number of parameters).
While large-scale models like Qwen2.5-VL-72B achieve strong performance,
our proposed LAD-Reasoner delivers comparable accuracy and even higher F1 with significantly fewer parameters (3B),
demonstrating superior efficiency.
APRIL-GAN and AnomalyGPT,
despite differing in scale,
both underperform in this trade-off space,
highlighting that scale alone is insufficient for complex anomaly reasoning.
Notably, both models lack the ability to generate reasoning traces,
suggesting limited support for interpretability and multi-step inference.
In contrast,
LAD-Reasoner is capable of producing coherent reasoning paths,
reinforcing the connection between structural reasoning ability and performance on logical anomaly detection.
These findings highlight the superiority of our model in both performance and reasoning capability, compared to these models.

\subsection{Reasoning Quality of GRPO}

\begin{figure*}[t]
  \centering
  \includegraphics[width=\linewidth]{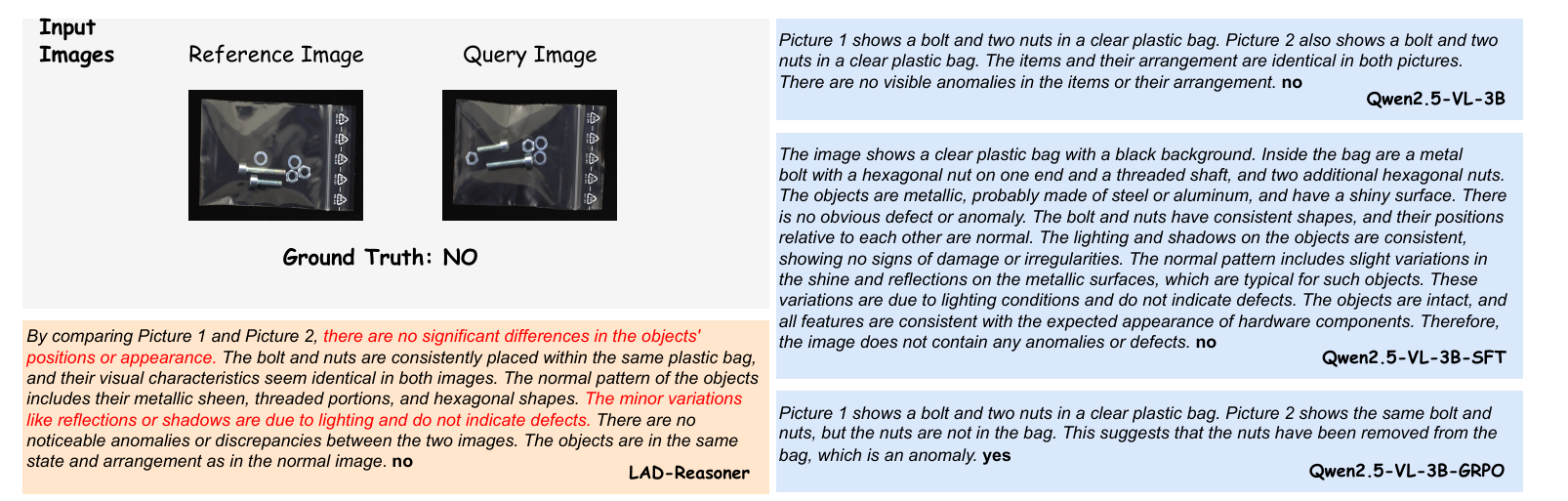}
  \includegraphics[width=\linewidth]{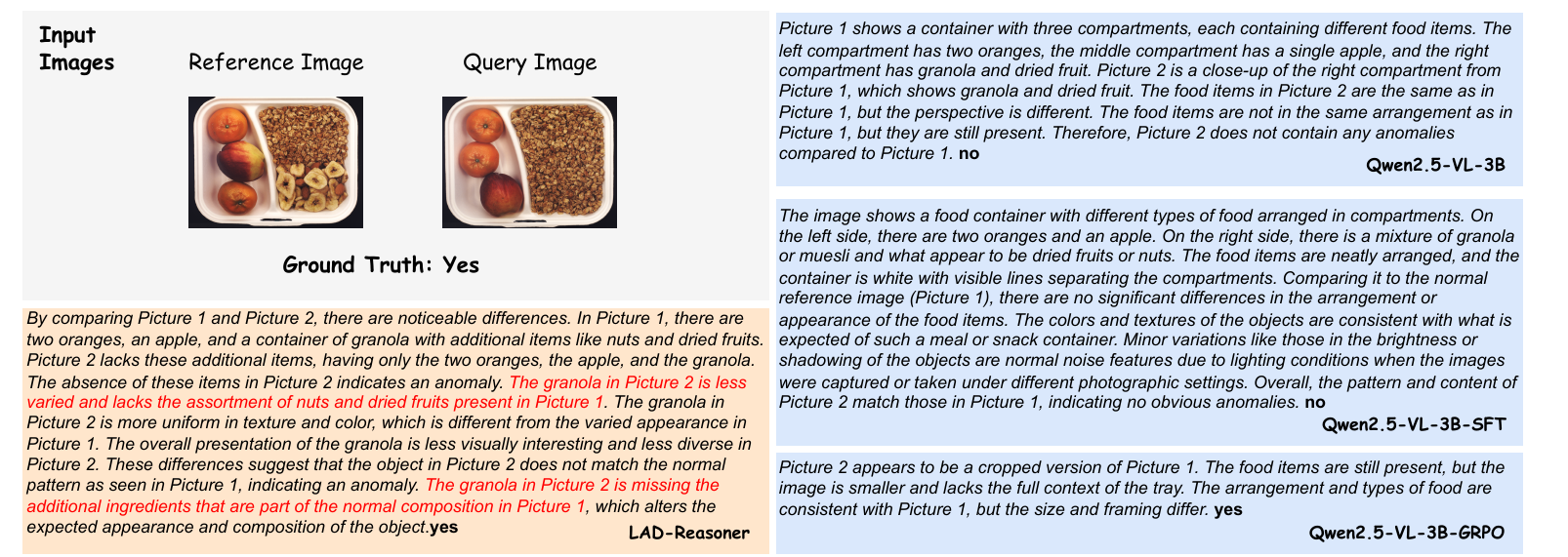}
  \caption{Comparison of the generated reasoning outputs across the base model, the SFT model, the GRPO model, and our proposed LAD-Reasoner.
LAD-Reasoner demonstrates improved perceptual details and a better reasoning process. Each response contains a thinking process (shown in \textit{italic}) followed by a binary decision (shown in \textbf{bold}). For clarity of presentation, the original \texttt{<think>}\texttt{<\textbackslash{}think>} and \texttt{<answer>}\texttt{<\textbackslash{}answer>} tags are omitted.}
  \Description{SFT Data Show.}
  \label{fig:ablation}
\end{figure*}

As shown in Figure~\ref{fig:ablation},
we compare model outputs on a normal case from the screw bag class,
aiming to assess whether each variant can reason about logical consistency beyond superficial visual similarity.
Qwen2.5-VL-3B and Qwen2.5-VL-3B-SFT correctly classify the case as normal,
but their explanations lack structural reasoning, focusing only on appearance.
In contrast, Qwen2.5-VL-3B-GRPO incorrectly predicts an anomaly due to missing items,
failing to recognize the context.
LAD-Reasoner not only makes the correct prediction but also provides a coherent explanation,
attributing minor visual differences to lighting while affirming spatial and compositional consistency.
We further evaluate an anomaly case from the breakfast box class.
Here, Qwen2.5-VL-3B and Qwen2.5-VL-3B-SFT miss the anomaly, citing framing or texture.
Qwen2.5-VL-3B-GRPO demonstrates better logical reasoning but lacks fine-grained visual sensitivity.
LAD-Reasoner alone accurately detects the missing ingredients and altered granola composition,
aligning these findings with the expected reference template.

Overall, SFT enhances perceptual detail, GRPO enables logical reasoning, and their integration in LAD-Reasoner achieves both accurate and interpretable anomaly understanding.

\section{Conclusion}
In this work,
we introduce Reasoning Logical Anomaly Detection (RLAD),
a novel task that requires generating a coherent reasoning process alongside the final anomaly judgment.
To address this challenge,
we present LAD-Reasoner, a multimodal framework trained with Supervised Fine-Tuning (SFT) and Group Relative Policy Optimization (GRPO).
Our approach jointly enhances detection accuracy and interpretability by producing structured,
human-readable rationales.
Extensive experiments on the MVTec LOCO AD dataset demonstrate that LAD-Reasoner achieves strong performance even with limited supervision,
and prove the tiny multimodal models can be good reasoners for RLAD.
Future work includes scaling to broader domains and incorporating richer data sources of reasoning supervision to further improve generalization and explainability.


\bibliographystyle{ACM-Reference-Format}
\bibliography{sample-base}

\end{document}